\documentclass[letterpaper, 10 pt, conference]{ieeeconf}  

\IEEEoverridecommandlockouts                              

\usepackage{cite}
\usepackage{graphics} 
\usepackage{epsfig} 
\usepackage{times} 
\usepackage{amsmath} 
\usepackage{amssymb}  
\usepackage{graphicx}
\usepackage[dvipsnames]{xcolor}
\usepackage{comment}   
\usepackage{caption}
\usepackage{subcaption}    
\usepackage[inkscapelatex=false]{svg}
\usepackage{xcolor}
\usepackage{xfrac}
\usepackage{pgfplots}
\usepackage{tikz, tikz-3dplot}
\usetikzlibrary{math}
\usetikzlibrary{tikzmark}
\usepackage{xparse}
\usepackage{float}
\usepackage{bbm}
\usepackage{url}
\usepackage{bm}
\usepackage{mathtools}
\usepackage{hyperref}
\usepackage{algorithm}
\usepackage{algpseudocode}

\pgfplotsset{compat=newest}

\NewDocumentCommand{\todo}{m}{\textcolor{red}{TODO: #1}}

\DeclarePairedDelimiter{\norm}{\|}{\|}

\DeclareMathOperator*{\argmin}{arg\,min}
 
\DeclareSymbolFont{boldoperators}{OT1}{cmr}{bx}{n}
\SetSymbolFont{boldoperators}{bold}{OT1}{cmr}{bx}{n}

\title{\LARGE \bf
Language-Grounded Control for Coordinated Robot Motion and Speech }

\author{Ravi Tejwani$^{1}$, Chengyuan Ma$^{1}$, Paco Gomez-Paz$^{2}$, Paolo Bonato$^{3}$ and H. Harry Asada$^{4}$,$~\IEEEmembership{Fellow,~IEEE}$%
\thanks{$^{1}$Ravi Tejwani and Chengyuan Ma are with Dept. of Electrical Engineering and Computer Science (EECS), Massachusetts Institute of Technology,
Cambridge, MA 02142 USA
{\tt\small \{tejwanir, macy404 \}@mit.edu}}%
\thanks{$^{2}$Paco Gomez-Paz is with Dept. of Mathematics, Massachusetts Institute of Technology,
Cambridge, MA 02142 USA
        {\tt\small pjgomez@@mit.edu}}%
\thanks{$^{3}$Paolo Bonato is with Harvard Medical School Department of Physical Medicine and Rehabilitation at Spaulding Rehabilitation Hospital, 
Charlestown, MA 02129 USA
        {\tt\small pbonato@mgh.harvard.edu}}%
\thanks{$^{4}$H. Harry Asada is with Dept. of Mechanical Engineering, Massachusetts Institute of Technology,
Cambridge, MA 02142 USA
        {\tt\small asada@mit.edu}}%
}

\begin{document}

\maketitle
\thispagestyle{empty}
\pagestyle{empty}

\begin{abstract}
Recent advancements have enabled human-robot collaboration through physical assistance and verbal guidance. However, limitations persist in coordinating robots' physical motions and speech in response to real-time changes in human behavior during collaborative contact tasks.
We first derive principles from analyzing physical therapists' movements and speech during patient exercises. These principles are translated into control objectives to: 1) guide users through trajectories, 2) control motion and speech pace to align completion times with varying user cooperation, and 3) dynamically paraphrase speech along the trajectory.
We then propose a Language Controller that synchronizes motion and speech, modulating both based on user cooperation. 
Experiments with 12 users show the Language Controller successfully aligns motion and speech compared to baselines. This provides a framework for fluent human-robot collaboration.
\end{abstract}


\section{INTRODUCTION}
Robots have been enabled to collaborate with humans by providing physical assistance as well as verbal guidance during collaborative tasks. Research on robots providing physical assistance has shown robots assisting with heavy lifting and materials handling in warehouses and factories \cite{muller2022}; handling payloads, reducing physical strain on human workers \cite{haninger2022towards}; and helping turn and lift patients, freeing up human nurses for other critical care tasks \cite{barhydt2023high}.
In addition to providing physical assistance, robots have also been used to provide verbal instructions and dialog interaction in human-robot collaboration \cite{selby2021teachbot, fong2003collaboration}. Using natural language capabilities, robots understand commands, ask clarifying questions, and provide guidance to human partners \cite{thomason2015learning}. Research suggests that language-enabled robots lead to higher perceived collaboration quality compared to silent robot partners \cite{szafir2021}. 

Human-robot collaboration remains limited by the lack of natural coordination between physical interactions and verbal communication. Simply combining robot's motions and its speech in parallel cannot achieve natural, fluent coordination. An open challenge remains to develop adaptive control frameworks that closely coordinate a robot's physical motions and speech utterances, dynamically modulating both based on real-time changes in human behavior.

We propose a Language Controller that dynamically aligns the robot's motion and speech under changing user cooperation so both end at the same time. It does so by varying admittance parameters, audio pace, and adaptive paraphrasing. 
The controller is inspired by principles derived from analysis of human-human physical interactions, specifically from an observational study of a physical therapist collaborating with a patient during therapeutic exercises at a rehabilitation center. From observations of the therapist's physical and verbal guidance, we identify core principles and translate them into formal control objectives for the controller (details in section ~\ref{sec:principles-objectives}).

We make the following contributions: 
\begin{enumerate}
    \item Formalize core principles to derive control objectives for natural human-robot collaboration -- adaptive pacing, aligned speech-motion timing, correlating speech complexity with motion speed -- from observations of human-human physical therapy interactions;
    \item Language Robot Controller from the derived control objectives in order to align the robot motion with the verbal speech in the human-robot interaction;
    \item Extensive human experiments that validate the controller and demonstrate its ability to align the pace of the robot motion with its speech.
\end{enumerate}




\section{RELATED WORK}
\subsection{Language Grounding for Robot Instructions}
Prior work has focused on grounding natural language instructions to enable robots to follow commands, including techniques for mapping instructions to internal representations and actions \cite{tellex2011understanding,matuszek2013learning,he2015deep}. Other efforts have targeted collaborative grounding of language between humans and robots for situated dialog and interactions \cite{chai2016collaborative,unhelkar2020decision}. While enabling planning and collaboration, integrating robot physical motions and speech grounded in real-time human responses still remains an open problem. Our work aims to address this gap.

\subsection{Variable Impedance and Admittance Control}
Research has explored variable impedance and admittance control for safer and adaptive human-robot interaction, including dynamic modulation based on cooperation \cite{dimeas2016online}, adaptive admittance using EEG feedback \cite{peternel2017adaptive}, and online impedance variation for performance/safety trade-offs \cite{gopinath2017human}. We incorporate admittance methods for compliant motion, but extend standard admittance frameworks by explicitly coupling the modulation of control parameters to the speech state. This ties motion control to verbal communication.



\subsection{Language and Motion Integration}
Recent works have combined language understanding with robotic planning and control, including mapping commands to executable specifications \cite{wang2021learning, 9197464} and leveraging implicit information to improve plan execution \cite{can2019learning, bisk2016natural}. However, physical motions of the robot and concurrent speech have not been integrated based on mutual understanding and real-time bidirectional communication. Our Language Controller addresses this by developing a control framework that coordinates motions and utterances grounded in human responses.

\section{PRINCIPLES AND CONTROL OBJECTIVES}
\label{sec:principles-objectives}
\begin{figure}[t!]
  \centering
  \includegraphics[width=0.4\textwidth]{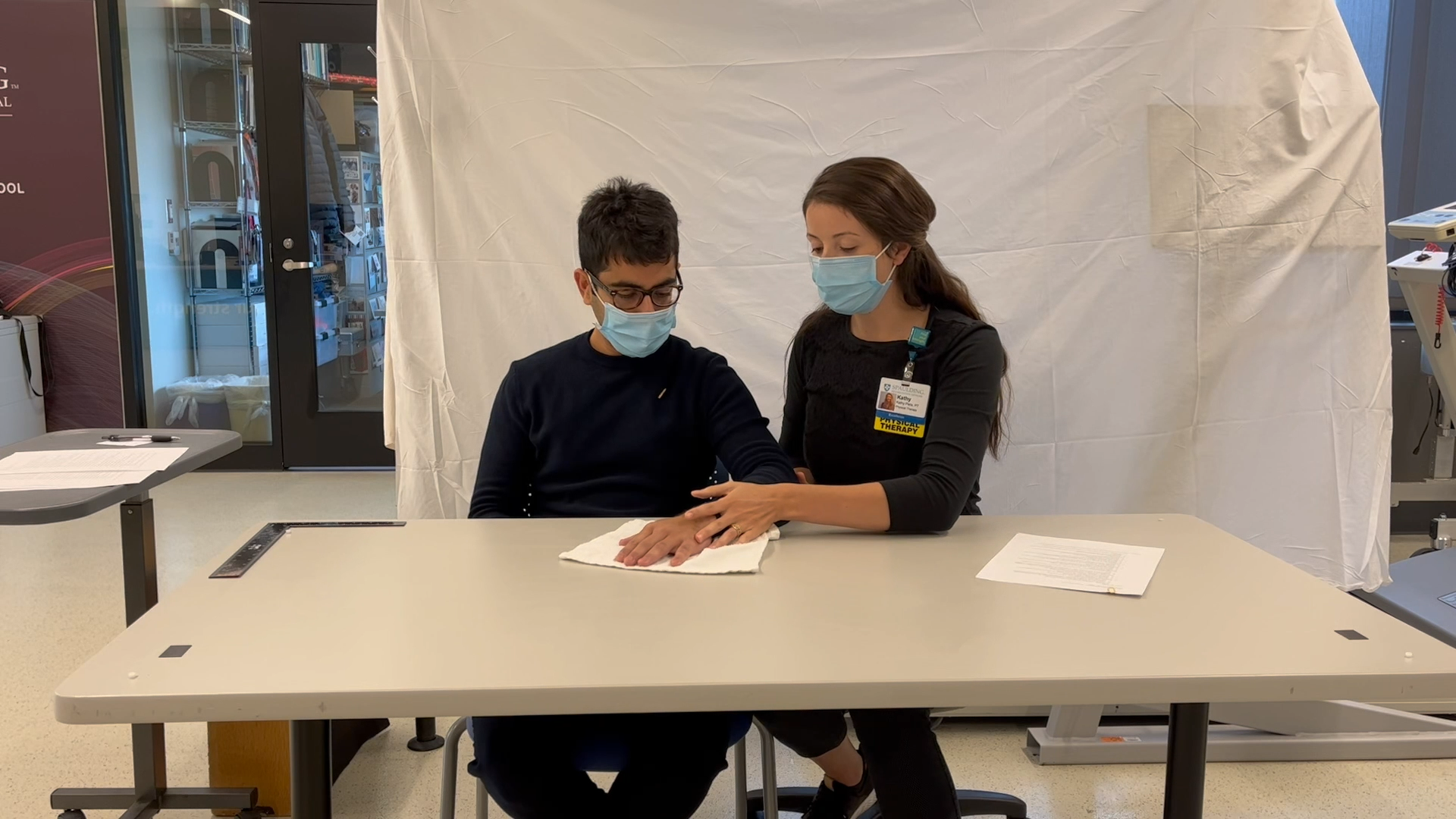}
  \caption{Physical therapist is seen performing 'shoulder external rotation' therapy on to the patient with varying levels of physical resistance. The physical motions and speech data was recorded across different sessions.\protect\footnotemark}
  \label{fig:therapy-session}
\end{figure}

Through the observational study of therapist-patient exercises at Spaulding Rehabilitation Hospital (Fig. ~\ref{fig:therapy-session}), we identified the following core principles:

\begin{itemize}
\item The therapist planned the trajectories for each session, demonstrating the path before starting an exercise;
\item The therapist adapted the pace of motions based on patient responses. When the patient struggled, she slowed down and gently guided them along the trajectory;
\item The therapist aligned her speech with physical actions. She began verbal guidance at the start of motions and finished speaking around the end;
\item Her speech rate and sentence length correlated with her physical motion speed - slower motions had slower, longer speech; faster motions had faster, shorter speech. 
\end{itemize}

We derived the following formal control objectives from these principles. The robot must:
\begin{enumerate}
\item Guide the user through a predefined trajectory while modulating its velocity in response to user cooperation. High cooperation (low resistance) must lead to faster motion and Low cooperation (high resistance) must lead to slower motion;
\item control the pace of its speech to maximize the alignment with its motion while adapting to varying user cooperation. We define alignment as the robot concluding its speech simultaneously as it completes motion.
\item paraphrase (choice of words) its speech dynamically along the trajectory, adapting to changing user cooperation. 
It must use shorter sentences under faster motion and longer sentences under slower motion.
\end{enumerate}

\footnotetext{Detailed therapy sessions videos can be seen at \url{https://language-playback-robot-controller.github.io/therapy-sessions/}}

\begin{figure*}
    \centering
    \includegraphics[width=0.8\textwidth]{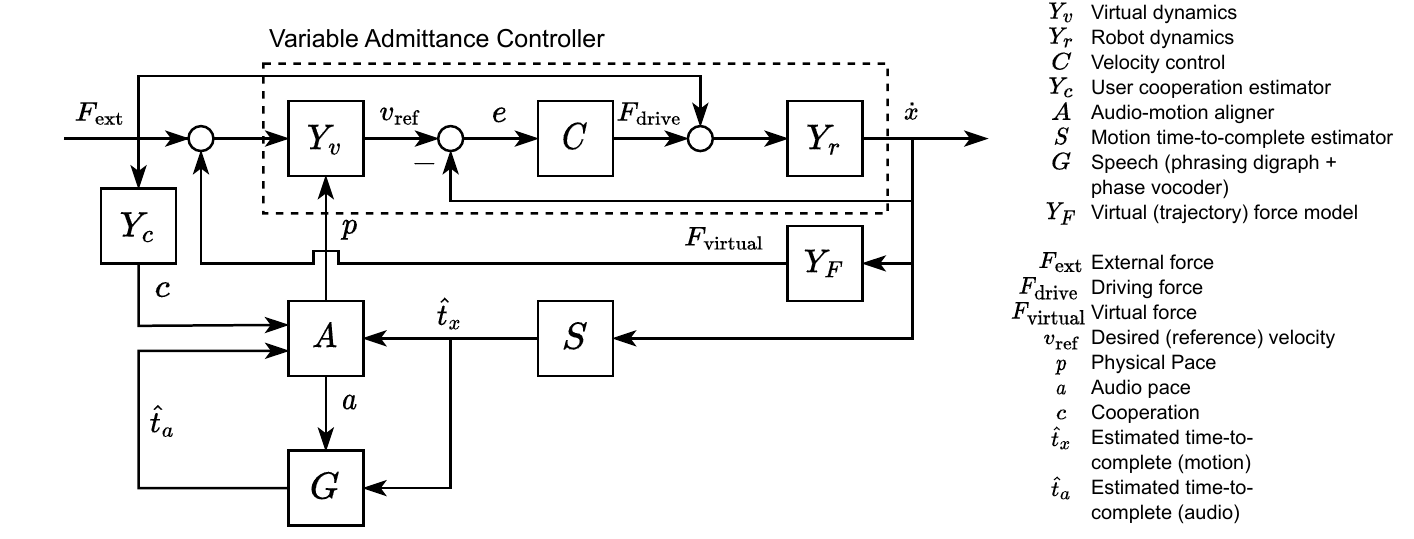}
    \caption{Control scheme of Language Controller. External force $F_{\text{ext}}$ (from user) and virtual force $F_{\text{virtual}}$ (based on position to guide user through a predefined trajectory) are passed to the virtual dynamics $Y_v$ to generate reference velocity $v_{\text{ref}}$ for velocity controller $C$. $C$ outputs a force that, together with $F_{\text{ext}}$, acts on robot dynamics $Y_r$. The resulting motion is given by $\dot x$. Motion time-to-completion estimator $S$ uses $\dot x$ to estimate the time-to-completion (ETC) of the trajectory $\hat{t}_x$. Speech module $G$ reports an ETC of the speech, $\hat{t}_a$, and cooperation model $Y_c$ computes cooperation $c$ from $F_{\text{ext}}$. $\hat{t}_x, \hat{t}_a$ and $c$ are fed to audio-motion aligner $A$ to update Physical Pace $p$ and Audio Pace $a$. Physical Pace $p$ changes the pace of the robot's motion by varying $Y_v$'s parameters. Audio Pace $a$ changes the pace of audio through a phase vocoder in $G$. }
    \label{fig:our-controller}
\end{figure*}

Building upon the outlined control objectives, we formalize the Language Controller.

\section{LANGUAGE CONTROLLER}
\subsection{Overview}

Language Controller (Fig.~\ref{fig:our-controller}) employs variable admittance control for robot motion and modulates the pace of motion and speech to maximize the alignment. The controller is designed to align the motion of the robot with its speech.
It does so by updating ``Physical Pace'' and ``Audio Pace'' from estimated time-to-completions for both the robot's trajectory and its speech. The controller dynamically updates the paces based on real-time user response. Furthermore, the controller incorporates adaptive paraphrasing to modulate the speech content. It traverses a phrase graph representation to select appropriate wording and phrase length that matches the expected duration of motion. 

\subsection{Admittance Model}
\label{sec:admittance}

Admittance and Impedance Control are the two primary control schemes used in human-robot interactions \cite{keemink2018admittance, hogan1984impedance}. Humans actively control their limbs and resist unexpected movements, which positions humans as an impedance, necessitating robots to be treated as an admittance. We thus use Admittance Control, which converts external force $F_{\text{ext}}$ into desired velocity $v_\text{ref}$ via a virtual dynamics model:
\begin{equation}
M_0 {\dot{v}_{\text{ref}}} + D_0 v_{\text{ref}} = F_{\text {ext}}.
\label{eq:virtual-dynamics}
\end{equation}
where $v_{\text{ref}}$ is the desired velocity.
To lead the user through a predefined trajectory $T$, we extend the virtual dynamics model above with a virtual force $F_{\text{virtual}}$,
\begin{equation}
M_0 {\dot{v}_{\text{ref}}} + D_0 v_{\text{ref}} = F_{\text {ext}} + F_{\text{virtual}}.
\label{eq:virtual-dynamics-driven}
\end{equation}
$F_{\text{virtual}}$ is dependent on the end effector position $x$ and its closest point on the trajectory ($x_d$), defined as follows,
\begin{equation}
    x_d = T(d), \text{ where } \textstyle d = \argmin_{0\le d \le 1} \norm{T(d) - x},
\end{equation}
where we see the trajectory $T$ as a directed curve $[0,1]\mapsto S$ where $S$ is the state space. $F_{\text{virtual}}$ consists of two parts:
\begin{align}
    F_{\text{virtual}} &= \underbrace{K(x_d - x)}_{F_{\text{guide}}} + 
    \underbrace{\textstyle \norm{F_{\text{propell}}} \mathbf{b}}_{F_{\text{propell}}}, \\
    \text{where }\mathbf{b}&= \lim_{d'\to d^+}\frac{T(d')-x_d}{\norm{T(d')-x_d}},
\end{align}
where $F_{\text{guide}}$ leads the user back on track if they deviate and  $F_{\text{propell}}$ leads the user to complete the trajectory. This achieves our first objective that our robot should lead the user through the trajectory. In frequency domain, \eqref{eq:virtual-dynamics-driven} can be expressed as 
\begin{equation}
    v_{\text{ref}} = A\cdot(F_{\text {ext}} + F_{\text{virtual}}) \text{ where } A=\frac{1}{M_0s + D_0}.
\end{equation}
We elaborate in \ref{sec:r-x} how to vary $A$ with Physical Pace $p$ to tune the behavior of this admittance model.

The virtual admittance model produces $v_{\text{ref}}$ that is fed to the velocity controller $C$, which produces a driving force $F_{\text{drive}}$ with actuators. The equation of motion of our robot is:
\begin{equation}
    M_{\text{robot}} \ddot{x} = F_{\text{ext}} + F_{\text{drive}} = F_{\text{ext}} + C(v_{\text{ref}} - \dot x).
    \label{eq:robot-motion}
\end{equation}




\subsection{Parameters}
\label{sec:sync}
Aligning the robot's motion and its speech requires us to be able to control their pace. In our controller, this is done by varying the \textbf{Audio Pace} $a$ and the \textbf{Physical Pace} $p$.

\textbf{Audio Pace ($a$)} is the pace at which the robot's speech is played. E.g., $a = 1.2$ means the audio is played $20\%$ faster than normal and $a = 0.8$ means playing $20\%$ slower. We pass $a$ to a \emph{Phase Vocoder}\cite{flanagan1966phase} which time-scales the prerecorded speech audio with Short-time Fourier Transform. We empirically constrain $a \in (0.6, 1.4)$ to avoid incongruity arising from over-stretching or shrinking of the speech audio.

\textbf{Physical Pace ($p$)} is a variable in our admittance controller. 
It can be thought of as a ``speed knob'' with which we vary the controller. 
We design $p$ to achieve the following effect: assuming constant $F_{\text{ext}}$, $p=1.2$ should cause our controller to complete a trajectory in $20\%$ less time than a fixed admittance controller following \eqref{eq:virtual-dynamics-driven} and likewise $p=0.8$ should cause it to run $20\%$. We empirically constrain $p\in(0.6, 1.4)$ so the admittance parameters do not deviate too much from their base values.
\textit{Physical Pace  ($p$) must not be confused with end effector velocity. }
\textit{The role of $p$ is not to directly modulate velocity, but to adjust the admittance parameters.} This ensures that the controller operates within the safety and compliance boundaries set by the admittance control framework. Directly multiplying the end effector velocity by $p$ could lead to unsafe conditions, as it would bypass these regulatory mechanisms. Therefore, $p$ should be understood as a rate constant that modifies the admittance parameters to indirectly influence end effector velocity, maintaining safety and compliance even under varying external forces. The formal definition of Physical Pace $p$ is in \ref{sec:r-x}.


\textbf{Estimated time-to-complete (ETC) for audio and speech} We define alignment of motion and speech as the two ending at the same time. Therefore, to modulate the paces, our controller naturally needs to continuously estimate when the motion and speech will end. Concretely, our controller computes the ETC for audio ($\hat{t}_a$) and motion ($\hat{t}_x$). $\hat{t}_a$ and $\hat{t}_x$ are computed \footnote{$\hat{t}_a$ is computed as the sum of audio length on a path constructed by repeated use of \eqref{eq:heuristic} from the current vertex in the phrasing graph (minus the duration played for the current audio ). $\hat{t}_x$ is computed assuming $F_{\text{ext}} = 0$ (i.e., fully cooperative user) with simulation of \eqref{eq:robot-motion} at 500Hz.
} \emph{under base pace $a=p=1$}\footnote{Both paces are defined relative to a base pace of $1$. $a=1$ means the audio is played at the recorded rate (free of distortions); $p=1$ means the trajectory is being run with the default / intended admittance parameters. We consider $1$ to be the most natural/ideal pace.}. 

\textbf{Computing Paces $p$ and $a$} We compute the ideal paces, $p^*$ and $a^*$, from the following optimization:
\begin{equation}
    \text{minimize } (p-p_{\text{natural}})^2+(a-a_{\text{natural}})^2, \text{ s.t. } \frac{\hat{t}_x}{p} = \frac{\hat{t}_a}{a},\label{eq:sync-opt}
\end{equation}
where we set $p_{\text{natural}} = a_{\text{natural}} = 1$. $\hat{t}_x / p$ and $\hat{t}_a / a$ are the ETC for motion and audio considering the current pace, and we equate them above to express the intent that both should end at the same time, i.e., aligned. The solution to \eqref{eq:sync-opt} is 
\begin{equation}
    p^* = \frac{s + 1}{s^2 + 1}, \quad a^* =  \frac{s^2 + s}{s^2 + 1} \text{ where } s = \frac{\hat{t}_x}{\hat{t}_a}.
\end{equation}
We further update $p$ and $a$ following the equation below:
\begin{align}
    \dot{p} = k_p(p^*-p),\quad \dot{a} = k_a(a^*-a). \label{eq:update-ra}
\end{align}
This control equation makes both paces converge exponentially to their optimal values\footnote{We do not directly set $p=p^*$ and $a=a^*$, which could lead to abrupt change of pace if $p^*$ and $a^*$ deviate from current values too much.}.

\textbf{User Cooperation} 
We define Cooperation ($0<c<1$) as 
\begin{equation}
    c(t) = 1 - \int_{0}^t \alpha^{t-\tau} \frac{\norm*{F_{\text{ext}}}}{\norm{F}_{\text{max}}} d\tau,
    \label{eq:resistance}
\end{equation}
where $\norm{F}_{\text{max}}$ is the maximum magnitude of resisting force and $\alpha$ is the decay factor. In practice, we apply a deadband filter to $F_{\text{ext}}$ first to filter out sensor noises and friction. 

We then extend \eqref{eq:update-ra} to 
\begin{align}
    \dot{p} = k_p(p^*-p),\quad \dot{a} = k_a(a^*-a) - k_c(1-c).
    \label{eq:update-ra-resist}
\end{align}
This allows our controller to slow down the speech when the user does not cooperate (high resistance) and resume ideal speech pace when the user fully cooperates (low resistance). \eqref{eq:update-ra-resist} achieves our second objective that the robot's motion and audio must be aligned under varying user cooperation.

\subsection{Varying Admittance Model with Physical Pace $p$}
\label{sec:r-x}
We now give a formal definition of the Physical Pace $p$ and integrate it into the virtual dynamics defined in \eqref{eq:virtual-dynamics-driven}. Let $v_{\text{ref}}^*(t)$ be the reference velocity generated by an admittance controller per \eqref{eq:virtual-dynamics-driven} (without $p$), and let $v_{\text{ref}}(t)$ be the reference velocity from our controller (with pace $p$). Assuming fixed $(F_\text{ext} + F_\text{virtual})$, we want
\begin{equation}
    v_{\text{ref}}(t) = pv_{\text{ref}}^*(pt)
    \label{eq:vref-pt-naive}
\end{equation}
which, after integrating both sides, implies that a controller with pace $p$ would reach a reference position in $1/p$ the time of a controller without $p$. E.g., when $p=2$, a controller without $p$ would take twice amount the time to reach the same position as a controller with $p$. More generally, pace $p$ varies with time. Let $p_t$ be the pace at time $t$, we want
\def\Rt{{\textstyle \int_{0}^t p_{\tau}\,d\tau}}
\begin{equation}
    v_{\text{ref}}(t) = p_t v_{\text{ref}}^*(\Phi), \text{ where } \Phi = \Rt.
    \label{eq:trajectory-playback-def}
\end{equation}
Here time $\Phi$ generalizes $pt$ for time-varying $p$ in \eqref{eq:vref-pt-naive}. The assumption of fixed $(F_\text{ext} + F_\text{virtual})$ in the definition is critical: \textit{$p$ generally is not a scale factor to the end effector velocity but instead modulates the admittance parameters.}
This approach is safer as $p$ effects the velocity only indirectly through the admittance model, which ensures the safety and compliance of our controller under varying forces.

We achieve \eqref{eq:trajectory-playback-def} with the variable admittance model: 
\begin{align}
    v_{\text{ref}} &= A(p_t)\cdot(F_{\text {ext}} + F_{\text{virtual}}), \\
    \text{ where } A(p_t) &= \frac{1}{\frac{1}{p_t^2}M_0s + \frac{1}{p_t}D_0 - \frac{\dot{p}_t}{p_t^3}M_0}.
 \label{eq:vac_control}
\end{align}

\begin{proof}
\def\Rt{{\textstyle \int_{0}^t p_{\tau}\,d\tau}}
Differentiate \eqref{eq:trajectory-playback-def} \& multiply both sides by $M_0$,
\begin{align}
    M_0\dot{v}_{\text{ref}}(t) &=  p_{t}^2 M_0\dot{v}_{\text{ref}}^*(\Phi)  +  \dot{p}_{t} M_0v_{\text{ref}}^*(\Phi).\\
    \shortintertext{Expanding $M_0\dot{v}_{\text{ref}}^*(t)$ by \eqref{eq:virtual-dynamics-driven}:}
    M_0 \dot{v}_{\text{ref}}(t) &=  p_{t}^2 (F_{\text{ext}}+F_{\text{virtual}}) - p_{t}^2 D_0 v_{\text{ref}}^*(\Phi) +  \dot{p}_{t} M_0 v_{\text{ref}}^*(\Phi)\nonumber.\\
    \shortintertext{Substitute $v_{\text{ref}}^*(\Phi) = v_{\text{ref}}(t)/p_t$ by \eqref{eq:trajectory-playback-def}}
    M_0 \dot{v}_{\text{ref}}(t)  &=  p_{t}^2 (F_{\text{ext}}+F_{\text{virtual}}) - (p_tD_0 - {\textstyle\frac{\dot{p}_t}{p_t}M_0})v_{\text{ref}}.
\end{align}
whose simplification then leads to \eqref{eq:vac_control}.
\end{proof}


\vspace{1em}
\noindent\emph{E. Passivity Guarantees}
\vspace{0.5em}
\setcounter{subsection}{5}

We show that our controller defined in \eqref{eq:vac_control} is passive. A passive system is a system that is constrained in such a way that it does not inject excessive energy or instability into the interaction \cite{wyatt1981energy}. Formally, a system is passive w.r.t. an input-output pair $(u(t), y(t))$ if and only if there exists a positive definite storage function $V$ over the system such that:
\begin{equation}
V(t) - V(0) \leq \int_{0}^{t} u(t)^T \cdot y(t) dt \quad \forall t > 0
\end{equation}

\textbf{Theorem 1:} \textit{Consider a controller of the form outlined in \eqref{eq:vac_control} operating with linear trajectory $T$. If $K$ is orthogonal and positive definite, $D_0$ is positive definite, and $p$ is lower-bounded by a positive value, then the system is passive with respect to the force-velocity ($F_{\text {ext}}, v_{\text{ref}}$) input-output pair.}

(Proof provided in the Online Supplementary Material \footnote{\label{supp} Online Supplemental Proof: \url{https://language-playback-robot-controller.github.io/proofs/}}.)

\begin{figure}
  \centering
  \includegraphics[width=0.35\textwidth]{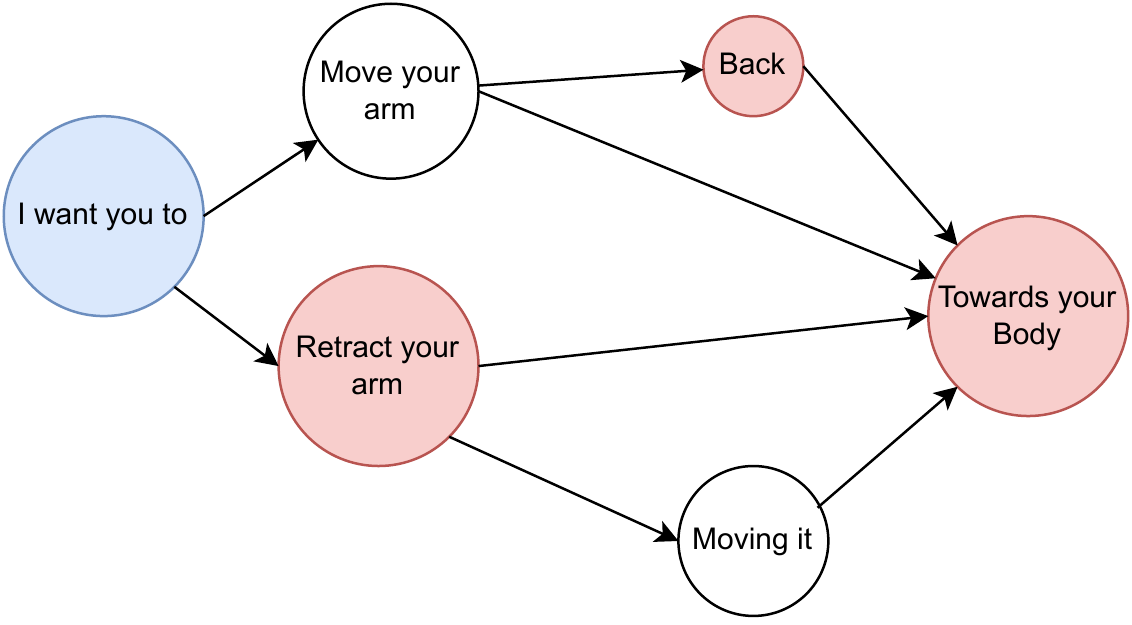}
  \caption{Example phrasing graph for our experiments where users were asked to retract their arms. 
  Paths ending at red nodes represent different phrasings of the instruction, e.g., ``I want you to retract your arm" (short) or ``I want you to move your arm back towards your body." (long)
  }
  \label{fig:phrasing-directed graph}
  \vspace{-2mm}
\end{figure}

\begin{figure*}
  \centering
  \includegraphics[width=0.90\textwidth]{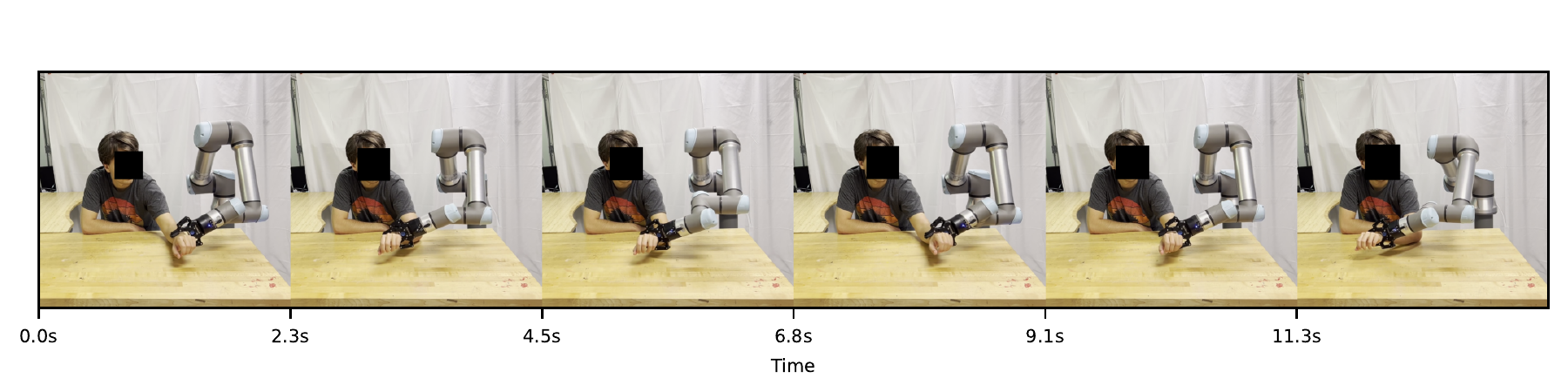}
  \caption{A user participant is seen interacting with the UR5 robot on a desired trajectory, which was inspired by the therapy session (``shoulder external rotation'') and predefined in our controller.\protect\footnotemark }
  \label{fig:session-frames}
  \vspace{-2.5mm}
\end{figure*}
\footnotetext{Results for all the users are available in our online appendix \url{https://language-playback-robot-controller.github.io/user-sessions/}.}
\begin{figure*}
    \centering
    \includegraphics[width=0.90\linewidth]{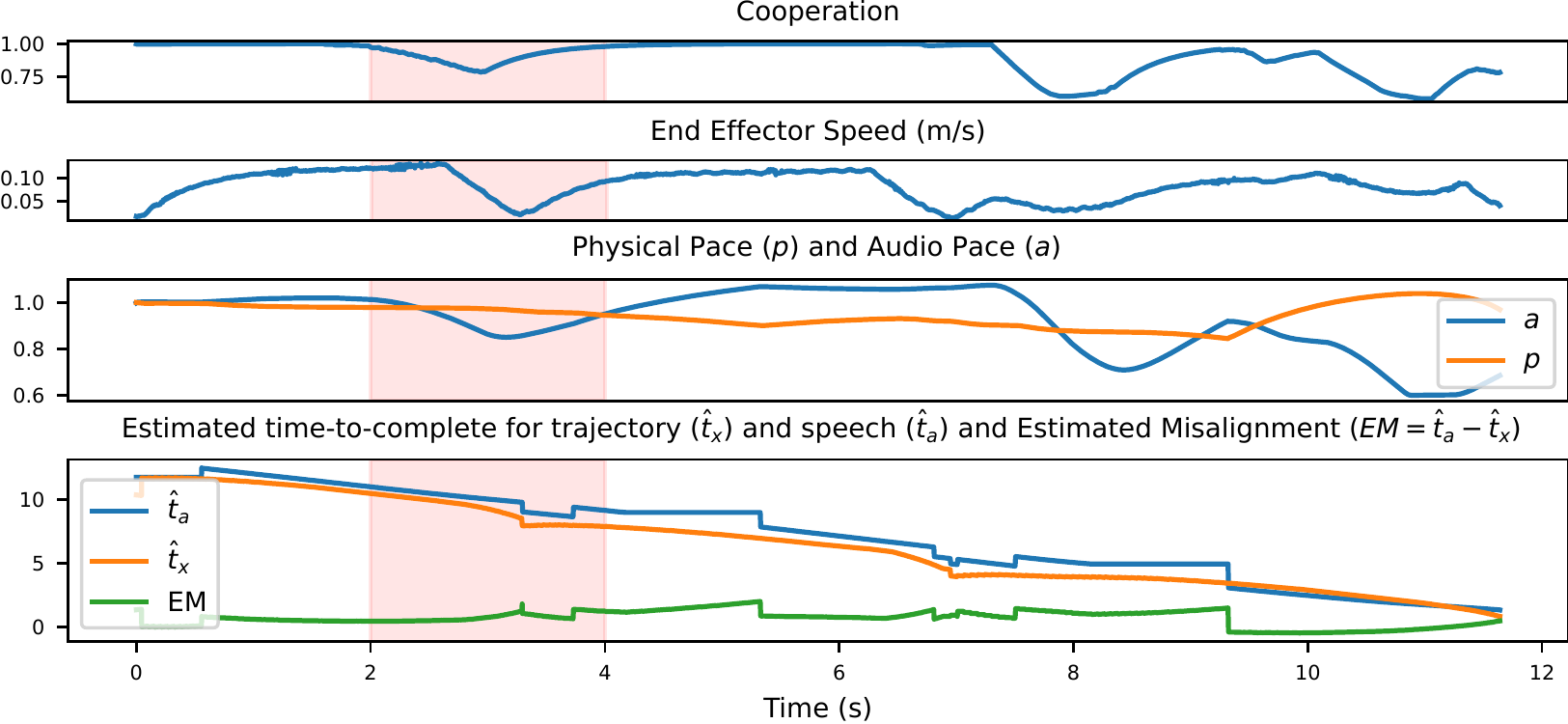}
    \caption{
    States and metrics of our controller from a single user session through a predefined intended trajectory. 
    At $t=3$ (\textcolor{red}{red shaded timespan}), cooperation dropped, causing end effector speed to decrease. This shows that our motion controller is compliant. Audio pace also dropped and returned back up at $t=4$, which shows our controller adapts the pace of the robot's speech to changing user cooperation.
    Throughout the session, our controller kept the estimated time-to-completion (ETC) of speech and trajectory close (the estimated misalignment is around 0). This demonstrates that our controller successfully aligns the robot's speech with its motion. 
    Around $t=7$, ETC for trajectory $\hat t_x$ dropped, making the controller select a shorter path on the phrasing graph, as shown by an abrupt drop in $\hat t_a$. However, the temporary dip in $\hat t_x$ was caused by inaccuracies of estimation, so the controller reverted to its original planned next phrase shortly after. As resistance dropped around $t=9$, the controller eventually paraphrased and chose the shorter path. The paraphrase shortened the ETC to less than the trajectory, so the controller re-aligns by reducing audio pace and increasing physical pace. Overall, this demonstrates that our controller is able to align the robot's speech with its motion while adapting to changing user cooperation.\protect\footnotemark[6]
    \protect\footnotemark}
    \label{fig:closeup-analysis}
  \vspace{-2mm}
\end{figure*}
\footnotetext{Step-like patterns for $\hat t_a$ around $t=5$ and $t=7$ is due to minor imperfection in our audio code when transitioning between different phrases at the time of the study. The steps should be lines of the same slope as the immediately preceding line and don't affect the correctness of our analysis.}

\subsection{Adaptively Paraphrasing the Robot Speech}
\label{sec:speech-directed graph}

To enable the robot to dynamically paraphrase its speech so the speech length matches that of the robot's motion (our third objective), we represent of speech as a \textbf{phrasing graph}. A phrasing graph is a Directed Acyclic Graph (DAG) where vertices denote sequences of words or phrases and a directed edge from vertex $u$ to $v$ denotes that the $v$'s phrase could follow $u$'s in speech. Phrasing graph captures the various alternative ways to express similar meanings (see Fig. \ref{fig:phrasing-directed graph}).

When the controller finishes saying the phrase on a vertex, it chooses the next vertex/phrase based on how long it expects the trajectory to last. Formally, it does so by choosing a next vertex $u$ from the graph, such that:
\begin{equation}
  u = \argmin_{\text{next node }u}\left|\hat{t}_x - \frac{\hat t_{\text{min}}(u) + \hat t_{\text{max}}(u)}{2}\right|
  \label{eq:heuristic}
\end{equation}
where $\hat{t}_x$ is the expected time-to-completion (ETC) of the trajectory (\ref{sec:sync}). $t_{\text{min}}(u), t_{\text{max}}(u)$ are the minimum \& maximum time-to-completion of speeches starting at vertex $u$\footnote{
$t_{\text{min}}(u)$ and $t_{\text{max}}(u)$ are pre-computed by storing at each vertex the expected time to say its phrase, and iterating in reverse topological order.}.

We remark that (a) (a) \eqref{eq:heuristic} causes the controller to select longer paraphrases when the user resists more. Higher resistance slows the motion, increasing trajectory time $\hat{t}_x$.
(b) the graph traversal does not depend on $a$ or $p$ to avoid compound effects\footnote{If we consider $p$ here, i.e., we choose next vertex $v$ that minimizes $|\hat{t}_x/p - \ldots|$ instead of $|\hat{t}_x - \ldots|$, a brief episode of low $p$ induces a longer speech which requires even lower $p$ to align -- a vicious cycle.}.
\begin{figure*}[h]
  \centering
  \includegraphics[width=0.8\linewidth]{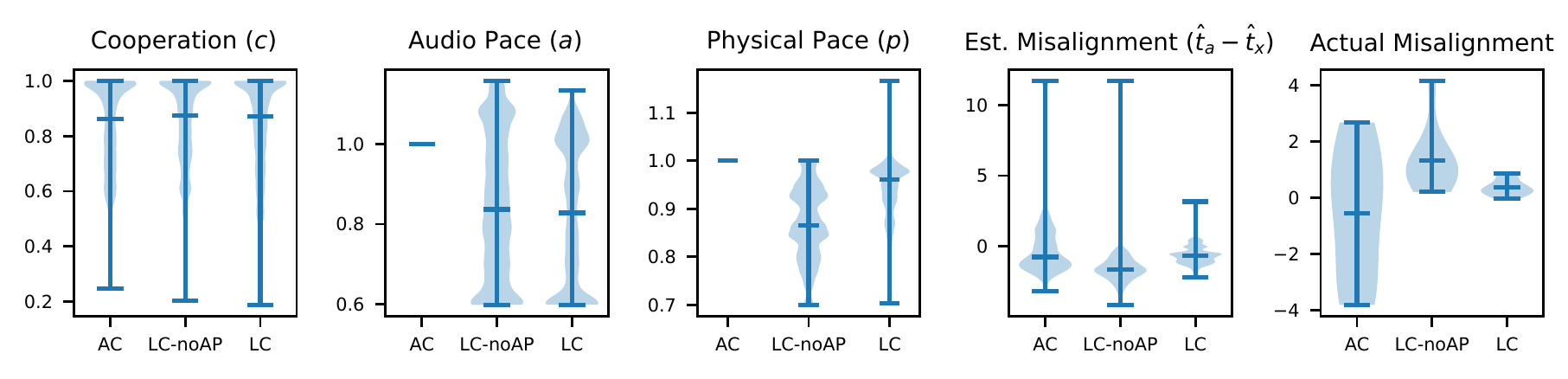}
  \caption{Violin plot of the distributions of metrics for 3 controllers across all 12 user sessions. Blue bars indicate min, median, and max of a distribution. Consistent distributions of cooperation for all controllers show that all controllers deliver similar physical experience to the users; On this ground, LC exhibits less audio-motion misalignment (both actual and predicted) than LC-noAR which exhibits significantly less variation in misalignment than the AC baseline. Adaptive paraphrasing allows LC to maintain a more natural speed of speech most of the time than LC-noAR, as shown by a more concentrated peak around the default rate of 1 in audio rate distribution. Overall, LC best aligns the robot's speech with motion\protect\footnotemark[6].}
  \label{fig:summary-plot}
  \vspace{-2mm}
\end{figure*}
\section{EXPERIMENTATION}

\subsection{User Study} 
The user study had 12 participants (7 males, 5 females, mean age 23). The UR5 robot \cite{ur5} guided users through a predefined trajectory demonstrated by the therapist. Users sat beside the robot, placed their hand on the desk. The robot guided their hand along the trajectory with speech instructions while users varied resistance arbitrarily.


\subsection{Control Schemes Evaluated and Compared} 
\begin{enumerate}
    \item \textbf{Admittance Controller with Decoupled Audio (\texttt{AC})}: Our baseline is a pure admittance controller with dynamics of \eqref{eq:virtual-dynamics-driven}. The audio is a single prerecorded audio file that starts simultaneously with the trajectory. Physical and Audio Paces are not modulated at all.
    \item \textbf{Language Controller without Adaptive Paraphrasing (\texttt{LC-noAP})}:  Our controller but without the ability to adaptively paraphrase the speech.
    \item \textbf{Language Controller (\texttt{LC})}: Our controller with adaptive paraphrasing by traversing a phrase graph (\ref{sec:speech-directed graph}).
\end{enumerate}

\subsection{Evaluation Metrics}
\begin{enumerate}
    \item \textbf{Cooperation} ($c$), defined in \eqref{eq:resistance}, quantifies user cooperation. $c$ should not significantly vary between control schemes. 
    \item \textbf{Audio Pace} ($a$) is the speed at which we deliver the speech audio. 
    A pace closer to 1 implies less distortion from audio processing and more natural speech.
    \item \textbf{Physical Pace} ($p$) is the state variable through which we vary the admittance parameters of our controller. $p$ closer to 1 implies the robot operating closer to its most natural admittance parameters. Always 1 for AC. 
    \item \textbf{Actual Misalignment} is defined as the difference between the audio and motion completion times. A positive value indicates the motion finished before the audio; A negative value indicates otherwise. A smaller absolute value suggests better audio-motion alignment.
    \item \textbf{Estimated Misalignment} ($\text{EM} = \hat t_{a} - \hat t_{x}$) is the real-time estimate of Actual Misalignment (AM). A smaller absolute value suggests better audio-motion alignment; EM differs from AM in that EM is a time series computed throughout the session whereas the AM is a scalar obtained after the session ends.
\end{enumerate}

\subsection{Analysis of Language Controller on a User Session}

We analyze our controller's behavior for one of the user sessions in Fig.~\ref{fig:closeup-analysis} and present a deep dive analysis. 
Around $t=3$, cooperation dropped, slowing end effector speed. Audio pace dropped then recovered, aligning with motion. Throughout, estimated time to completion (ETC) of speech and trajectory were kept close, showing speech-motion alignment. Around $t=7$, a temporary inaccurate ETC drop caused paraphrasing to a shorter phrase path, which was committed once resistance dropped around $t=9$.
Overall, Language Controller adapts speech pace to follow motion changes from user cooperation variations, while paraphrasing aligns speech ETC to trajectory ETC.

\subsection{Comparing Controllers across User Sessions}

We compare all three control schemes across all users and present the results in Fig.~\ref{fig:summary-plot}. 
AC exhibited broad misalignment (-4 to 3 seconds). Given short session length, this indicates notable lack of alignment. Adding pace control (LC-noAP) reduced misalignment, and adaptive paraphrasing (LC) further improved to under 1 second. Both LC finish speech after motion.
LC controllers showed bimodal audio pace distribution - a peak around base rate 1, another at lower end -- corresponding to two behaviors: slowing speech during high resistance to align with slower motion, and aiming for natural rate without/low resistance. LC has more concentrated peak at 1, as adaptive paraphrasing matches speech content length to motion, enabling natural rate delivery.

The experiments show that our controller met all of our defined objectives. It guided the users through the intended trajectory while adapting to the changing user resistance, and it controlled the pace and content of the robot's speech to maximize its alignment with the robot's motion.

\section{LIMITATIONS}
\begin{enumerate}
    \item We employed variable admittance controller for motion control. But the idea generalizes to other controllers;
    \item Speech length can be varied with filler words and pauses, which our controller does not yet implement;
    \item For richer interactions, large language models could automatically create phrasing graphs, beyond our therapist-recorded interactions; 
    \item Our controller assumes prerecorded audio phrases. An extension is using text-to-speech to generate audio from the phrases;
    \item Formalized for physical therapy, Language Controller principles can be extend to manufacturing, with robots assisting in lifting, handling, and assembly; and space exploration, helping astronauts recover from falls.
\end{enumerate}

\section{CONCLUSION}
We present a Language Controller that aligns robot motion and speech for human collaboration, formalizing principles from analyzing therapist interactions into control objectives. This enables compliant trajectories while modulated speech rate and paraphrasing align content length. Experiments validate motion-speech alignment of the controller over baselines. Future work expands its capabilities and applies to assistive manufacturing, healthcare, and space explorations.


\bibliographystyle{IEEEtran}
\bibliography{references} 

\end{document}